\documentclass[fleqn,10pt]{wlscirep}

\usepackage{graphicx}
\usepackage{url}
\usepackage{amsmath}

\usepackage{subfigure}
\usepackage{rotating}
\usepackage{algorithmic}
\usepackage{algorithm}
\usepackage{color}
\usepackage{soul}
\usepackage{subfigure}
\usepackage{array}
\usepackage{booktabs}
\usepackage{multirow}

\usepackage{amssymb}

\title{Towards automatic pulmonary nodule management in lung cancer screening with deep learning}

\author[1,2*]{Francesco Ciompi}
\author[1]{Kaman Chung}
\author[1]{Sarah J. van Riel}
\author[1]{Arnaud Arindra Adiyoso Setio}
\author[1]{Paul K. Gerke}
\author[1]{Colin Jacobs}
\author[1]{Ernst Th. Scholten}
\author[1]{Cornelia Schaefer-Prokop}
\author[3]{Mathilde M. W. Wille}
\author[4]{Alfonso Marchian\`o}
\author[4]{Ugo Pastorino}
\author[1]{Mathias Prokop}
\author[1]{Bram van Ginneken}
\affil[1]{Diagnostic Image Analysis Group, Radboud University Medical Center, Nijmegen, The Netherlands}
\affil[2]{Department of Pathology, Radboud University Medical Center, Nijmegen, The Netherlands}
\affil[3]{Dept. of Respiratory Medicine, Gentofte Hospital, Copenhagen, Denmark}
\affil[4]{Fondazione IRCCS Istituto Nazionale dei Tumori, Milano, Italy}

\affil[*]{francesco.ciompi@radboudumc.nl}

\begin{abstract}
The introduction of lung cancer screening programs will produce an unprecedented amount of chest CT scans in the near future, which radiologists will have to read in order to decide on a patient follow-up strategy.
According to the current guidelines, the workup of screen-detected nodules strongly relies on nodule size and nodule type.
In this paper, we present a deep learning system based on multi-stream multi-scale convolutional networks, which automatically classifies all nodule types relevant for nodule workup.
The system processes raw CT data containing a nodule without the need for any additional information such as nodule segmentation or nodule size and learns a representation of 3D data by analyzing an arbitrary number of 2D views of a given nodule.
The deep learning system was trained with data from the Italian MILD screening trial and validated on an independent set of data from the Danish DLCST screening trial.
We analyze the advantage of processing nodules at multiple scales with a multi-stream convolutional network architecture, and we show that the proposed deep learning system achieves performance at classifying nodule type that surpasses the one of classical machine learning approaches and is within the inter-observer variability among four experienced human observers.
\end{abstract}
\begin{document}

\flushbottom
\maketitle
%
%
\thispagestyle{empty}


\section*{Introduction}
The American National Lung Screening Trial (NLST) \cite{Aber11} demonstrated a lung cancer mortality reduction of 20\% by screening of heavy smokers using low-dose Computed Tomography (CT), compared with screening using chest X-rays. 
Motivated by this positive result and subsequent recommendations of the U.S. Preventive Services Task Force \cite{Koni13}, lung cancer screening is now being implemented in the U.S., where high-risk subjects will receive a yearly low-dose CT scan with the aim of (1) checking for the presence of nodules detectable in chest CT and (2) following-up on nodules detected in previous screening sessions.
As a consequence, an unprecedented amount of CT scans will be produced, which radiologists will have to read in order to check for the presence of nodules and decide on nodule workup.
In this context, (semi-) automatic computer-aided diagnosis (CAD) systems\cite{Mess10a,Jaco14,Seti15a,Seti16} for detection and analysis of pulmonary nodules can make the scan reading procedure efficient and cost effective.

Once a nodule has been detected, the main question radiologists have to answer is: \emph{what to do next?}
In order to address this question, the Lung CT Reporting And Data System (Lung-RADS) has been recently proposed, with the aim of defining a clear procedure to decide on patient follow-up strategy based on nodule-specific characteristics such as nodule \emph{type}, size and growth.
Lung-RADS guidelines also refer to the PanCan model \cite{McWi13}, which estimates the malignancy probability of a pulmonary nodule detected in a baseline scan (i.e., during the first screening session) based on patient data and nodule characteristics.
In both Lung-RADS guidelines and the PanCan model, the key characteristic to define nodule follow-up management is \emph{nodule type}.

\begin{figure}[t]
\centering
{\includegraphics[width=.8\linewidth]{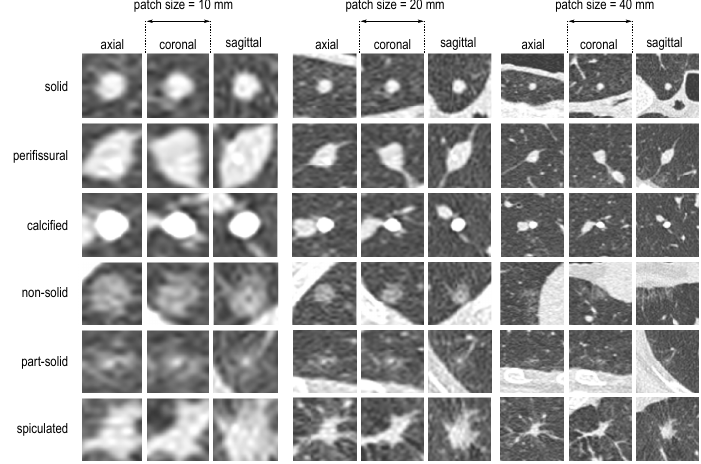}}
\caption{Examples of triplets of patches for different nodule types in axial, coronal and sagittal views. Each triplet is depicted using three different patch sizes, namely 10 mm, 20 mm and 40 mm.
\label{fig:nodules}}
\end{figure}

Pulmonary nodules can be categorized into four main categories, namely \emph{solid, non-solid, part-solid} and \emph{calcified} nodules (see Figure \ref{fig:nodules}). Solid nodules are characterized by a homogeneous texture, a well-defined shape and an intensity above -450 Hounsfield Units (HU) on CT. 
Two sub-categories of nodules with the density of solid nodules can be considered, namely \emph{perifissural} nodules \cite{Hoop12}, i.e., lymph nodes (benign lesions) that are attached or close to a fissure, and \emph{spiculated} nodules, which appear as solid lesions with characteristic spicules on the surface, often considered as an indicator of malignancy.
\emph{Non-Solid} nodules have an intensity on CT lower than solid nodules (in the range between -750 and -300 HU), also referred to as ground glass opacities.
\emph{Part-Solid} nodules contain both a non-solid and a solid part, the latter normally referred to as the \emph{solid core}.
Compared with solid nodules, non-solid and in particular part-solid nodules occur less frequent but  have a higher frequency of being malignant lesions \cite{Hens02}.
Finally, \emph{calcified} nodules are characterized by a high intensity and a well-defined rounded shape on CT.
Completely calcified nodules represent benign lesions.

In Lung-RADS, the workup for pulmonary nodules is mainly defined by nodule type and nodule size.
However, presence of imaging findings that increase the suspicion of lung cancer, such as spiculation, can modify the workup.
In the PanCan model, spiculation is a parameter that together with nodule type, nodule size and patient data contribute to the estimation of the malignancy probability of a nodule.
Furthermore, completely calcified and perifissural nodules are given a malignancy probability equal to zero.
In a scenario in which CAD systems are used to automate the lung cancer screening workflow from nodule detection to automatic report with decision on nodule workup, it is necessary to solve the problem of automatic classification of nodule type.
In this context, the classes that have to be considered are: ($i$) solid, ($ii$) non-solid, ($iii$) part-solid, ($iv$) calcified, ($v$) perifissural and ($vi$) spiculated nodules.

Although the general characteristics of nodule types can be easily defined, recent studies \cite{Riel15,Jaco15} have shown that there is a substantial inter- and intra-observer variability among radiologists at classifying nodule type.
In this context, researchers have addressed the problem of automatic classification of nodule type in CT scans by (1) designing a problem-specific descriptor of lung nodule and (2) training a classification model to automatically predict nodule type.
In \cite{Jaco15}, nodules were classified as solid, non-solid and part-solid.
A nodule descriptor was designed based on information on volume, mass and intensity of the nodule, and a kNN classifier was applied, but the used features strongly rely on the result of a nodule segmentation algorithm, whose optimal settings also depend on nodule type.
The authors propose to solve this problem by first running the algorithm multiple times using different segmentation settings in order to extract features and then classify nodule type. In practice, this strategy hampers the applicability of such a system to an optimized scan reading scenario.
In \cite{Fara10}, the SIFT descriptor was used to classify nodules as juxta, well circumscribed, pleural-tail and vascularized, and a feature matching strategy was used for classification purposes.
Despite the good performance reported, the considered categories are not relevant for nodule management according to current guidelines.
A descriptor specifically tailored for lung nodule analysis was introduced in \cite{Ciom14c}, which was used to assess presence of spiculation in detected solid nodules \cite{Ciom15} and to classify nodules as perifissural \cite{Ciom15a}.
Although this approach could be extended to other nodule types, it strongly relies on the estimation of nodule size in order to define the proper \emph{scale} to analyze data.

Scale is an important factor to consider in automatic nodule type classification.
As an example, discriminating a pure solid nodule from a perifissural nodule involves the detection of the fissure, which on a 2D view of the nodule can be differentiated from a vessel only if a sufficiently large region surrounding the nodule is considered (see Figure \ref{fig:nodules}).
On the other hand, discriminating non-solid from part-solid nodules strongly relies on the presence of a solid core, which can consist of a tiny part of the lesion that can only be clearly detected on a small scale.

In recent years, the advent of \emph{deep learning} \cite{Lecu15,Schm15a} has emerged as a powerful alternative to designing ad-hoc descriptors for pattern recognition applications by using deep neural networks, which can learn a representation of data from the raw data itself.
The most used incarnation of deep neural networks are convolutional networks\cite{Lecu98,Kriz12,Lecu15}, a supervised learning algorithm particularly suited to solve problems of classification of natural images\cite{Kriz12,Serm14,Szeg14}, which has recently been applied to some applications in chest CT analysis\cite{Ginn15,Ciom15a,Seti16,Tara16,Anth16}.

In this paper, we address the problem of automatic nodule classification by introducing three main contributions.
For the first time, we propose a single system that classifies all nodule types relevant for patient management in lung cancer screening according to the Lung-RADS assessment categories and the PanCan malignancy probability model, namely solid, non-solid, part-solid, calcified, perifissural and spiculated nodules.
Differently from what has been done in previous work, we design a classification framework based on Convolutional Networks (ConvNets) \cite{Schm15a,Lecu98,Kriz12}.
In particular, we propose a \emph{multi-stream multi-scale} architecture in which ConvNets simultaneously process multiple triplets of 2D views of a nodule at multiple scales and compute the probability for the nodule to belong to each one of the six considered nodule types.
The proposed approach does not require nodule segmentation or the estimation of nodule size.
Inspired by recent work\cite{Ciom15a,Seti16}, we formulate the analysis of a nodule as a combination of 2D patches.
Relying on the experimental results of Setio et al.\cite{Seti16}, which showed that performance increase by increasing the number of analyzed patches, we go beyond a limited number of patches by introducing a novel approach to extract an arbitrary number of 2D views from a nodule.
We trained the deep learning system using data from 943 patients and 1,352 nodules from the Multicentric Italian Lung Detection (MILD) trial\cite{Past12} and we validated the trained system using independent data from 468 patients and 639 nodules from the Danish Lung Cancer Screening Trial (DLCST) \cite{Pede09}.
Furthermore, in order to compare the performance of our deep learning architecture versus classical approaches of patch classification, we trained a linear support vector machines classifier to classify both features based on the raw intensity of nodules and features learned from raw data via an unsupervised learning approach.
Finally, in order to compare the performance of our method versus human performance, we designed an observer study in which four observers, including experienced radiologists, classified a subset of 162 nodules extracted from the test set.
We show that the proposed system achieves performance that surpasses classical patch classification approaches and is comparable with the inter-observer variability among human observers.

\begin{table}[t]
\scriptsize
\renewcommand{\arraystretch}{1.3}
\centering
\noindent\begin{tabular}{*{6}{|c}|}
\hline
 & \multicolumn{4}{c|}{\textbf{MILD} (943 patients)} & \multicolumn{1}{c|}{\textbf{DLCST} (468 patients)}\\
\hline
			& Training nodules & $N$ & Training samples &  Validation nodules & Test nodules $test_{ALL}$ / $test_{OBS}$\\			
\hline
Solid & 694 & 8 & 88,832 & 232 & 382 / 27\\
Calcified & 233 & 22 & 82,016 & 78 & 58 / 27\\
Part-solid & 63 & 80 & 80,640 & 21 & 37 / 27\\
Non-solid & 152 & 33 & 80,256 & 50 & 87 / 27\\
Perifissural & 181 & 28 & 81,088 & 62 & 48 / 27\\
Spiculated & 29 & 167 & 77,488 & 10 & 27 / 27\\
\hline
\textbf{Total} & 1,352 & -- & 490,320 & 453 & 639 / 162\\
\hline
\end{tabular}
\caption{Detailed number of nodules and samples in the training, validation and test sets. The MILD dataset is used for training and validation purposes, the DLCST dataset is used for testing purposes. In the test set, the number of nodules per class randomly selected to design the observer study is reported. The number of class-specific planes per nodule used to extract training data ($N$, see also Figure \ref{fig:deeplearning}) is indicated for each nodule type. The number of used patients from MILD and DLCST are also indicated.}
\label{tab:data}
\end{table}

\section*{Results}

\subsection*{Training data}
We trained the deep learning system using data from the Multicentric Italian Lung Detection (MILD) trial \cite{Past12}.
For this purpose, we considered all baseline CT scans from the MILD trial.
The study was approved by the Institutional review board of Fondazione IRCCS Istituto Nazionale Tumori di Milano, and the written informed consent was waived for the retrospective examination of the analyzed data.
For all patients, non contrast-enhanced low-dose CT scans were acquired using a 16-detector row CT system, with section collimation 16 $\times$ 0.75 mm.
Images were reconstructed using a sharp kernel (Siemens B50 kernel, Siemens Medical Solutions) with a slice thickness of 1.0 mm.
Nodules were detected and annotated based on the following procedure.
All CT scans were first read by a workstation (CIRRUS Lung Screening, Diagnostic Image Analysis Group, Radboudumc, Nijmegen, Netherlands) with automatic nodule detection (CAD) tools integrated.
Two medical students, trained by a radiology research in detecting pulmonary nodules, either accepted or rejected CAD marks and labeled nodules as one of the considered nodule types.
Accepted nodules were segmented using the algorithm presented in\cite{Kuhn06}, which is implemented in CIRRUS Lung Screening.
The students manually adjusted parameters to obtain the best possible nodule segmentation, which allowed to compute the equivalent diameter of the lesion.
Nodules with label disagreement were reviewed by a thoracic radiologist (ES) with more than 20 years of experience in reading chest CT scans.
Nodules with label agreement were further reviewed by two radiology researchers (SvR, KC) independently.
From the set of annotated nodules, we removed all cases with a diameter smaller than 4 mm, which is considered as an irrelevant finding in lung cancer screening\cite{Aber11}.
The final set of data consisted of 1,805 nodules from 943 subjects  (see Table \ref{tab:data}), which were split into two non-overlapping sets: a \emph{training} set (1,352 nodules), used to train the deep learning system and a \emph{validation} set (453 nodules), used to monitor the performance of the system during training.

In the development of the proposed deep learning system, we defined a nodule data \emph{sample} as a set of triplets of patches (axial, coronal and sagittal view), where each triplet was used to feed three streams of convolutional network (details on data preprocessing, system design and training are detailed in the Methods section).
For training purposes, several different samples were extracted from the same nodule by rotating triplets around the center of mass and by using techniques of data augmentation at patch level.
In this way, $\approx$ 0.5M training samples were extracted and used to train the deep learning system.
In our experiments, we investigated the performance of the system when data at different scales were considered.
For this purpose, we extracted nodule data with patches of size 10 mm, 20 mm and 40 mm, which represent 3 different scales.
We built and trained three network architectures where one scale (40 mm), two scales (20 mm, 40 mm) and three scales (10 mm, 20 mm, 40 mm) were processed, and compared the performance of the three networks with both classical patch classification approaches based on machine learning and human performance.

\subsection*{Test data}
The performance of the trained deep learning system was assessed on data from the Danish Lung Cancer Screening Trial (DLCST) \cite{Pede09}.
In particular, we used the subset of data used in a study recently published by the DLCST research group\cite{Wink15a}, where the authors also describe the procedure used to annotate nodule types.
The DLCST was approved by the ethics committe of Copenhagen County and fully funded by the Danish Ministry of Interior and Health.
Approval of data management in the trial was obtained from the Danish Data Protection Agency.
The trial is registered with ClinicalTrials.gov (NCT00496977).
All participants provided written informed consent.
Non contrast-enhanced low-dose CT scans were acquired using a multi-slice CT system (16-row Philips Mx 8000, Philips Medical Systems) with section collimation 16 $\times$ 0.75 mm.
Images were reconstructed using a sharp kernel (kernel D) with a slice thickness of 1.0 mm.
From the initial data set, we removed nodules with a diameter smaller than 4 mm, as done for data from the MILD trial, and discarded scans with incomplete or corrupted data (e.g., missing slices).
Finally, we obtained a set $test_{ALL}$ of 639 nodules from 468 subjects (see Table \ref{tab:data}), which we used for testing purposes.

\subsection*{Observer study}
In order to compare the deep learning system with human performance, we selected a subset of nodules from the set $test_{ALL}$ and asked three observers to label nodule type.
For this purpose, we built a dataset by including all spiculated nodules in $test_{ALL}$ (27 nodules) and the same number of nodules randomly selected from the other classes.
Therefore, a dataset $test_{OBS}$ of 162 nodules was built for the observer study.
Two chest radiologists (ES, CSP) with more than 20 years of experience reading chest CT and a radiology researcher (KC) were involved in the observer study.
Readers independently labeled nodule types.
Nodules were shown at locations indicated by annotations provided by the DLCST trial, and readers had the possibility to either label the nodule as belonging to one of the six considered categories, or label it as \emph{not a nodule}.
For evaluation purposes, we considered annotations made by the three observers involved in this study as well as annotations coming from the DLCST trial, which we considered as an additional observer.
In the rest of the paper we will refer to annotations coming from these four different sources as observers O$_1$, O$_2$, O$_3$ and O$_4$ (where O$_4$ indicates the DLCST annotations).

\begin{table}[t]
\scriptsize
\renewcommand{\arraystretch}{1.3}
\centering
\begin{tabular}{| c | c | c | c | c | c | c | c |}
\hline
 & \multicolumn{4}{c|}{Observers} & \multicolumn{3}{c|}{Computer}\\ 
\hline
 & O1 & O2 & O3 & O4 & 1 scale & 2 scales & 3 scales \\ 
\hline\hline
O1 & -- & 0.59 (0.51--0.68) & 0.65 (0.57--0.74) & 0.68 (0.60--0.76) & 0.63 (0.54--0.72) & 0.64 (0.55--0.73) & 0.65 (0.57--0.74) \\ 
O2 & 0.59 (0.51--0.68) & -- & 0.71 (0.63--0.79) & 0.66 (0.58--0.75) & 0.55 (0.45--0.64) & 0.54 (0.45--0.64) & 0.58 (0.49--0.67) \\ 
O3 & 0.65 (0.57--0.74) & 0.71 (0.63--0.79) & -- & 0.75 (0.67--0.82) & 0.56 (0.47--0.65) & 0.57 (0.48--0.66) & 0.61 (0.52--0.70) \\ 
O4 & 0.68 (0.60--0.76) & 0.66 (0.58--0.75) & 0.75 (0.67--0.82) & -- & 0.62 (0.53--0.70) & 0.64 (0.55--0.73) & 0.67 (0.59--0.75) \\ 
\hline
\end{tabular}
\caption{Cohen $\kappa$ statistics with 95\% confidence intervals for agreement between computer and observers. O$_i$ indicates the i-th observer. Results for automatic classification using deep learning systems with different numbers of scales are reported.}
\label{tab:kappa}
\end{table}

\begin{table}[t]
\scriptsize
\renewcommand{\arraystretch}{1.3}
\centering
\begin{tabular}{| c | c | c | c | c | c | c | c | c |}
\hline
 & Accuracy & $F_{Solid}$ & $F_{Calcified}$ & $F_{Part-solid}$ & $F_{Non-solid}$ & $F_{Perifissural}$ & $F_{Spiculated}$ & $F_{Not-a-nodule}$ \\ 
\hline
\hline
O1 vs. Computer (3 scales) & 71.5\%  & 60.8\%  & 88.4\%  & 66.7\%  & 86.3\%  & 62.2\%  & 71.4\%   & --\\ 
O2 vs. Computer (3 scales) & 66.2\%  & 62.6\%  & 82.4\%  & 47.8\%  & 72.7\%  & 80.0\%  & 56.4\%   & --\\ 
O3 vs. Computer (3 scales) & 67.7\%  & 56.8\%  & 85.1\%  & 59.1\%  & 78.3\%  & 75.6\%  & 60.9\%   & --\\ 
O4 vs. Computer (3 scales) & 72.8\%  & 64.2\%  & 88.9\%  & 71.7\%  & 80.0\%  & 77.3\%  & 62.7\%   & --\\ 
\hline
Average & 69.6\% & 61.1\% & 86.2\% & 61.3\% & 79.3\% & 73.8\% & 62.9\% & --\\
\hline
\hline
O1 vs. O2 & 66.0\%  & 52.7\%  & 84.0\%  & 51.3\%  & 79.2\%  & 63.6\%  & 83.3\%  & 50.0\%   \\ 
O1 vs. O3 & 71.0\%  & 55.0\%  & 87.0\%  & 66.7\%  & 80.0\%  & 81.5\%  & 74.4\%  & 40.0\%   \\ 
O1 vs. O4 & 72.8\%  & 64.8\%  & 90.9\%  & 66.7\%  & 71.7\%  & 75.5\%  & 89.4\%  & 0.0\%   \\ 
O2 vs. O3 & 76.5\%  & 74.7\%  & 88.9\%  & 61.5\%  & 81.0\%  & 77.3\%  & 75.7\%  & 66.7\%   \\ 
O2 vs. O4 & 72.2\%  & 64.4\%  & 88.5\%  & 70.8\%  & 71.1\%  & 79.1\%  & 73.2\%  & 0.0\%   \\ 
O3 vs. O4 & 79.0\%  & 68.4\%  & 95.8\%  & 71.1\%  & 80.9\%  & 90.6\%  & 79.2\%  & 0.0\%   \\ 
\hline
Average & 72.9\% & 63.3\% & 89.2\% & 64.7\% & 77.3\% & 77.9\% & 79.2\% & 26.1\%\\
\hline
\end{tabular}
\caption{Nodule classification performance in terms of accuracy and F-measure per nodule type. Results for each pair of human observer O$_i$ vs. O$_j$ and for observers versus the computer on the $test_{OBS}$ dataset (167 nodules) are reported. Averages of measures across observers and across computer-observers are also indicated. The additional class ``not a nodule'' is added to observers since they could exclude nodules during the observer study.}
\label{tab:comparison}
\end{table}

\begin{table}[t]
\scriptsize
\renewcommand{\arraystretch}{1.3}
\centering
\begin{tabular}{| c | c | c | c | c | c | c | c |}
\hline
 & Accuracy & $F_{Solid}$ & $F_{Calcified}$ & $F_{Part-solid}$ & $F_{Non-solid}$ & $F_{Perifissural}$ & $F_{Spiculated}$\\ 
\hline
Intensity features + SVM & 27.0\%  & 4.1\%  & 60.2\%  & 0.0\%  & 35.4\%  & 26.7\%  & 32.5\%  \\ 
\hline
Unsupervised features + SVM & 39.9\%  & 38.4\%  & 32.0\%  & 49.4\%  & 59.2\%  & 16.9\%  & 39.5\%  \\ 
\hline
ConvNets 1 scale & 78.0\% & 84.4\% & 82.4\% & 54.5\% & 84.4\% & 57.5\% & 37.8\%\\
\hline
ConvNets 2 scales & 79.2\% & 85.6\% & 84.9\% & 52.3\% & 87.8\% & 63.4\% & 36.8\%\\
\hline
ConvNets 3 scales & 79.5\% & 85.6\% & 85.7\% & 52.2\% & 87.4\% & 68.2\% & 43.4\%\\
\hline
\end{tabular}
\caption{Comparison of classification performance on the $test_{ALL}$ set (639 nodules) in terms of accuracy and F-measure when the considered methods are: (1) features based on pixel intensity of patches and linear SVM classifier, (2) features learned from raw nodule patches using the unsupervised learning approach proposed in \cite{Coat11} and linear SVM classifier, (3) the proposed deep learning approach using ConvNets working at 1, 2 and 3 scales. In these experiments, annotations from DLCST radiologists (O$_4$) are considered as the reference standard.}
\label{tab:comparison_machine_learning}
\end{table}

\begin{figure}
\centering
{\includegraphics[width=\linewidth]{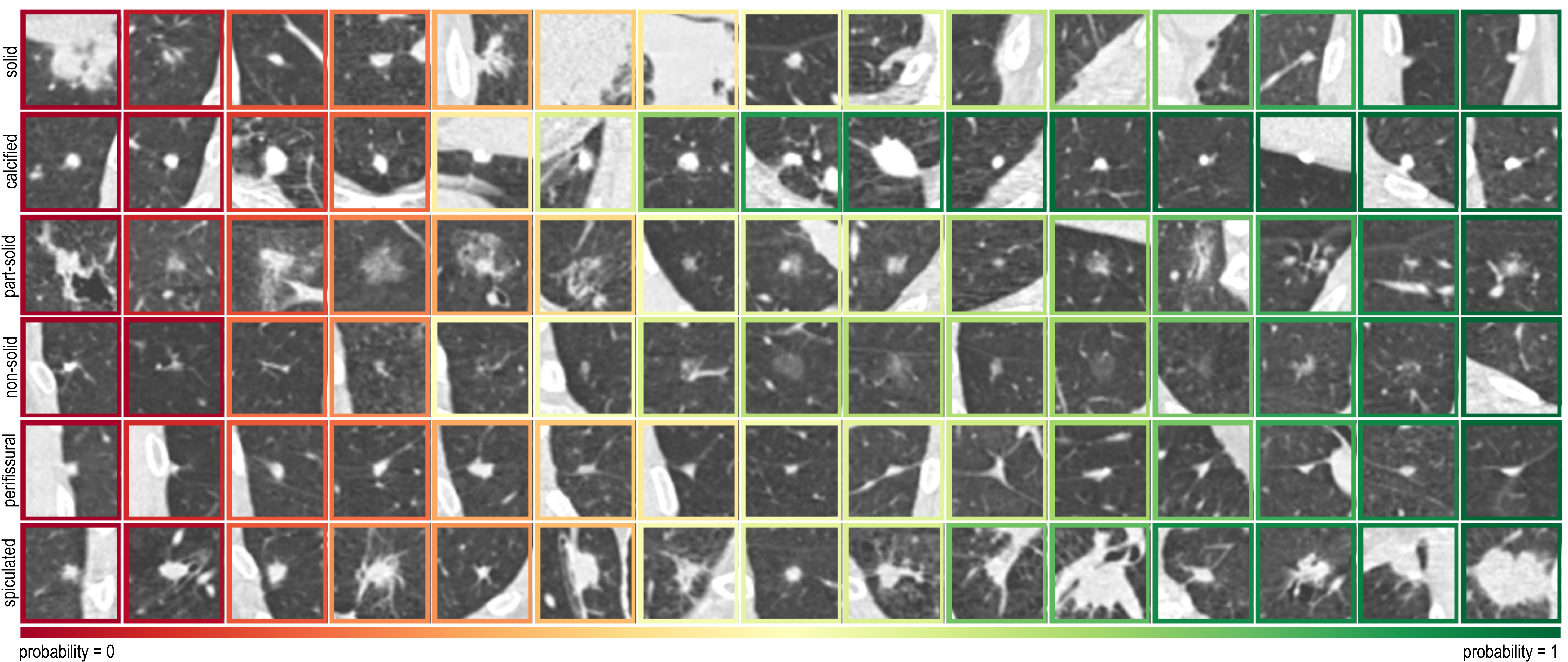}}
\caption{Examples of classified nodules from the test set (DLCST). Each row depicts nodules from one class as labeled in the DLCST trial, and nodules are sorted from left to right based on the probability given by the (3-scale) deep learning system. Examples with low probability (on the left) are a-typical cases of each nodule type, while a high probability (on the right) is given to typical examples of each nodule type.
\label{fig:sorted_nodules}}
\end{figure}

\begin{figure}
\centering
{\includegraphics[width=1.0\linewidth]{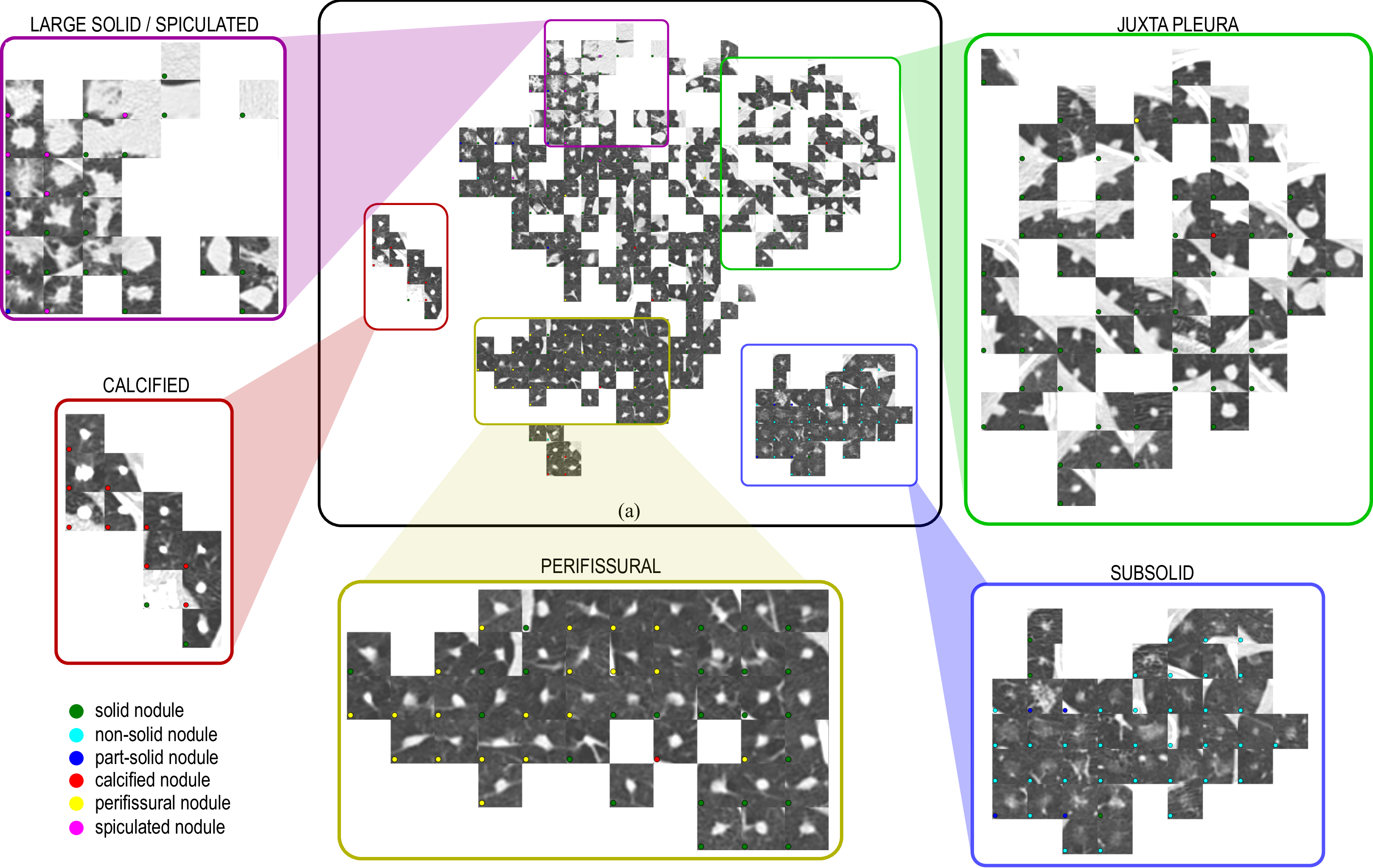}}
\caption{In (a), multidimensional scaling of nodules in the test set using the t-SNE algorithm. Close nodules have similar characteristics.
In (a), clusters of similar nodules are highlighted and grouped with different boxes.
A zoomed-in version of each cluster is also shown and a representative name is given based on their appearance.
The nodule label assigned in the DLCTS trial is also reported as a coloured dot for each nodule patch (see legend for nodule types).
\label{fig:tsne}}
\end{figure}

\subsection*{Evaluation}
After training, all nodules in $test_{ALL}$ were classified using the trained deep learning system.
In order to compute the computer-observer agreement, we compared the results from the computer with the nodule type given by each observer independently in the $test_{OBS}$ set.
Furthermore, we computed the inter-observer agreement by considering all possible pairs of observers O$_i$ vs. O$_j$ ($i, j = 1,\dots,4$, $i\neq j$).
In this case, since observers were given the possibility of labeling a given nodule as ``not a nodule'', the additional class \emph{not a nodule} is considered to assess the inter-observer variability.
The results in terms of $\kappa$ value are reported in Table \ref{tab:kappa}, when all pairs of observers and the results from the three deep learning architectures working with different scales are considered.
It can be noted that human observers have a moderate to substantial agreement, with $\kappa$ between 0.59 and 0.75, and that the deep learning system achieves a variability in the same range of human observers, with a level of agreement that increases with the number of scales used for nodule classification.
When the 3-scale architecture is considered, the $\kappa$ value between the computer and each observer under test is between 0.58 and 0.67 and in half of the cases, it is higher than the agreement between the observer under test and at least one of the other observers.

We also evaluated the classification performance of the best performing network, namely the one working with 3 scales, in terms of accuracy and per-class F-measure and compared it with human performance (Table \ref{tab:comparison}).
It is worth noting that the average performance among human observers are comparable with the average performance between the computer and observers, with an average accuracy of 72.9\% versus 69.6\% respectively.
A similar trend can be observed for all the other classification parameters.

Furthermore, we used the $test_{ALL}$ dataset to compare the performance of the proposed deep learning system with two \emph{classical} approaches where a linear Support Vector Machines (SVM) classifier was trained in a supervised fashion using features extracted from 2D nodule patches. In the first approach, features based on the raw pixel intensity of 2D patches were used (\emph{intensity features}). In the second approach, features were not engineered but learned from raw data via an unsupervised learning approach using the K-Means algorithm (\emph{unsupervised features}), as proposed in \cite{Coat11}.
Details on the design of these two additional experiments are given in the Methods section.
The proposed approach based on deep learning, together with these two approaches based on classical machine learning, covers a scenario where the problem of nodule classification is tackled by (1) manually defining features based on raw image data and use them to train a classifier, (2) learning features from raw data in an unsupervised fashion and use them to train a classifier, (3) learning a hierarchical representation of nodules from raw data, using convolutional networks trained end-to-end.
The results of the comparison are reported in Table \ref{tab:comparison_machine_learning}, where the gradual improvement from using intensity-based features and SVM to a 3-scale approach based on deep learning can be observed both in terms of accuracy and F-measure.

In Figure \ref{fig:sorted_nodules}, examples of nodule type classification are depicted, grouped based on labels provided by the DLCST trial.
For each nodule type, nodules classified by the deep learning system are ordered by increasing probability.
As a consequence, atypical examples for each nodules type can be found on the left  side of the figure, while typical examples can be found on the right side of the figure.

\subsection*{Discussion}
The deep learning system produces a score by classifying an internal representation learned from raw data.
In order to get insights on the kind of features learned by the network, we extracted an embedded representation of each nodule and applied multidimensional scaling to project the embedded representation onto a bidimensional plane.
For this purpose, we applied the t-Distributed Stochastic Neighbor Embedding (t-SNE) algorithm \cite{Maat08} to the output of the last fully-connected layer of the network.
In this way, each nodule is represented by a feature vector of 256 values.
The result of the multidimensional projection is depicted in Figure \ref{fig:tsne}, where close nodules have a similar representation in the network.
Clearly defined clusters of nodules with similar characteristics can be identified.
Examples are clusters of large solid nodules, calcified or perifissural nodules, but also groups of nodules of a particular class that was not used in this study, namely juxtapleural nodules.

One of the clusters in the t-SNE representation shows a direct association between large solid nodules and spiculated nodules.
Based on training data, the system implicitly learns that large solid nodules are likely to be spiculated nodules.
This effect can be observed in the quantitative evaluation reported in Table \ref{tab:comparison}, where spiculation has an F-measure of 62.7\% when the system is compared with O$_4$ on the subset of 162 nodules, while it decreases to an F-measure of 43.4\% when all nodules are considered.
The reduction in precision observed in the second experiment is therefore related to the presence of more large solid nodules that are misclassified as spiculated.
This suboptimal behavior of the system can be compensated by increasing the amount of spiculated nodules in the training set, for example by including follow-up cases.
Nevertheless, in the clinical context of lung cancer screening, labeling large solid nodules as spiculated may not hamper the nodule workup, since large solid nodules without spiculation are also considered as suspicious lesions.

The values of precision and recall per nodule type when the $test_{ALL}$ set is classified with the 3-scale network are reported in Table \ref{tab:precision_recall}.
We can observe that the system tends to classify solid, calcified and non-solid nodules with high performance.
As a consequence, since nodule type distributions are skewed (see Table \ref{tab:data}), the overall accuracy for $test_{ALL}$ is higher than for $test_{OBS}$.
The low value of precision and recall for part-solid and spiculated nodules in $test_{ALL}$ corroborates what is observed for $test_{OBS}$ and can be compensated in the future by adding more training samples for underrepresented classes, therefore increasing the variability of nodule appearance in the learning procedure.

The performance of the system is within the inter-observer variability.
This corroborates the effectiveness of the system at classifying nodules and also indicates that even experienced radiologists do not fully agree on nodule types.
The concept of nodule type has been coined by radiologists, who have to differentiate opacities in CT scans according to their appearance and, most importantly, to their frequency of malignancy.
The fact that there is no complete agreement among experienced radiologists implies that no gold standard for nodule type classification can be made, and that there will always be doubtful cases even in the training set.
In this context, the range of variability within the one among humans reached by the proposed system makes it the first suitable system to be integrated in workstations for automatic analysis of CT scans in lung cancer screening.

\begin{table}[t]
\scriptsize
\renewcommand{\arraystretch}{1.3}
\centering
\begin{tabular}{| c | c | c | c | c | c | c |}
\hline
 & Solid & Calcified & Part-solid & Non-solid & Perifissural & Spiculated \\ 
\hline
Precision & 89.2 & 88.9 & 43.6 & 87.4 & 78.4 & 32.7  \\ 
\hline
Recall & 82.2 & 82.8 & 64.9 & 87.4 & 60.4 & 64.3  \\ 
\hline
\end{tabular}
\caption{Precision and recall values for the 3-scale deep learning system tested the $test_{ALL}$ set.}
\label{tab:precision_recall}
\end{table}

\begin{figure}[t]
\centering
{\includegraphics[width=\linewidth]{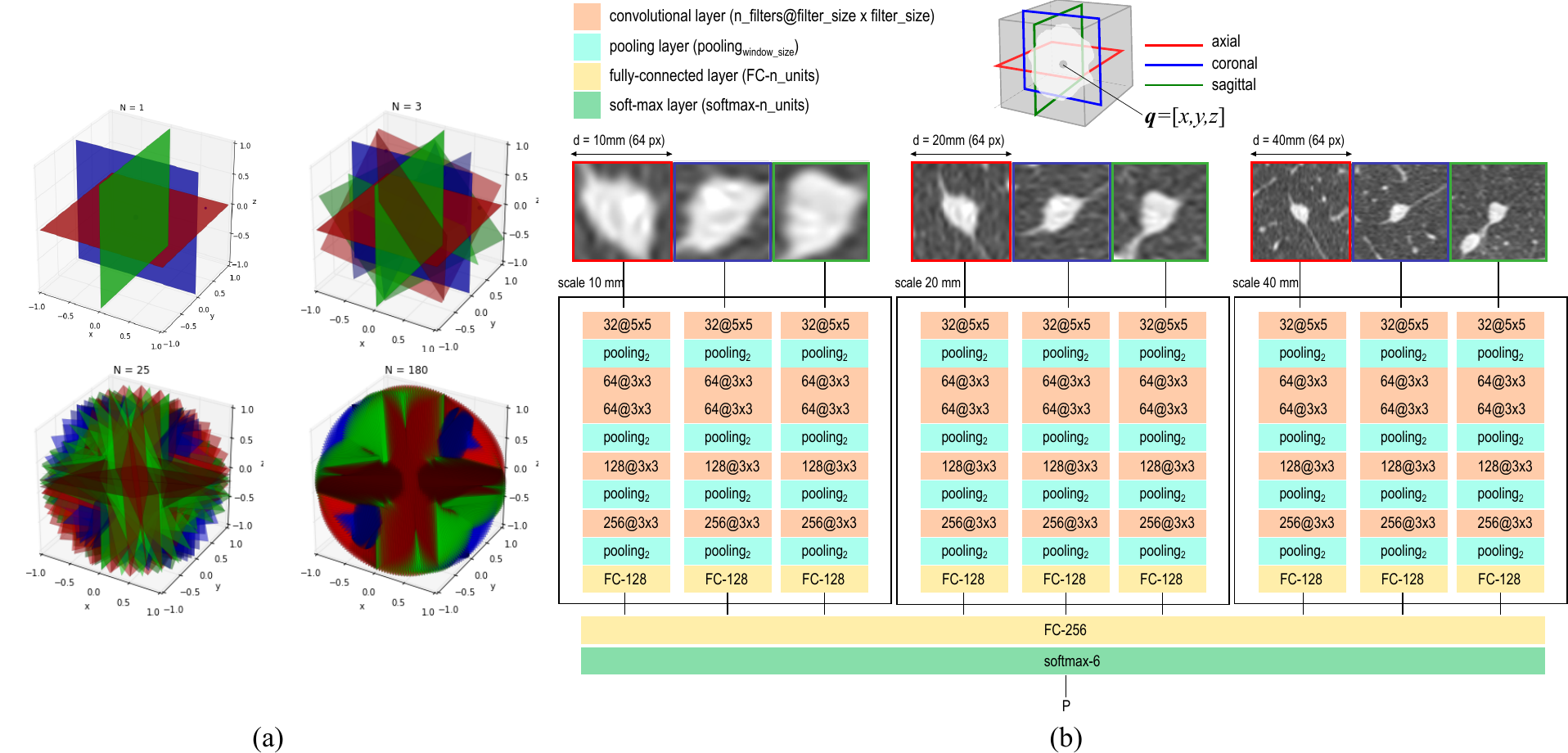}}
\caption{(a) Examples of triplets of nodules extracted by varying the parameter $N$. (b) Examples of pyramidal triplets of patches used to feed the proposed deep learning systems. The system consists of three groups of three streams, one for each considered scale (namely 10 mm, 20 mm and 40 mm for patch size). Convolutional layers, max-pooling layers, fully-connected layers and one soft-max layer are the building blocks of the proposed network. The last fully-connected layer with 256 neurons serves as a combiner of the three sets of three streams, and a 6-value probability vector is generated as output.
\label{fig:deeplearning}}
\end{figure}

\section*{Methods}\label{sec:methods}
The input of the proposed framework is a chest CT scan and the position $\mathbf{q} = [x, y, z]$ of the nodule (e.g., its center of mass) to classify.
The output of the system is the probability for the nodule to belong to each one of the six considered classes.
The framework is based on convolutional networks (ConvNet), which process input samples via a ``multi-stream multi-scale'' architecture (see Fig. \ref{fig:deeplearning}).
We define an input \emph{sample} as a triplets of 2D patches obtained by intersecting the 3D domain of the nodule with triplets of orthogonal planes, and crop triplets of patches at different resolutions.
Therefore, an input sample to feed the deep learning system is given by three triplets of patches from the same nodule (see Fig. \ref{fig:deeplearning}).
Each step of the proposed framework is detailed in next sections.  

\subsection*{Generation of triplets of 2D patches}
Let us define a triplet of orthogonal planes $\mathcal{T}_n = \{\Psi_n, \Omega_h, \Phi_n\}$ passing through the point $\mathbf{q}$ and an angle $\theta_n = \frac{(n-1)\pi}{2N}$ (\mbox{$n=1,\dots,N$}), which defines the rotation of each plane of $\mathcal{T}_n$ with respect to the axes $x, y, z$.
In this way, $\mathcal{T}_1$ is the triplet of planes that define the default axial, coronal and sagittal views of a CT scan, and any other triplet $\mathcal{T}_n$ is obtained by sequentially rotating the triplet with respect to the $x$, the $y$ and the $z$ axis by an angle $\theta_n$.
Rotating all the planes by the same angle guarantees that orthogonal planes are always obtained.
Examples of triplets for several values of $N$ are depicted in Figure \ref{fig:deeplearning}(a), where the axial, coronal and sagittal planes are represented in different colors.

The intersection of a triplet of planes and a CT scan generates 2D views of the nodule of interest.
From each intersection, we generate triplets of 2D patches by cropping a square area of size $d$ centered on $\mathbf{q}$.
Increasing the value of $N$ allows to increase the number of extracted patches per nodule, which also increases the coverage of the volume of a nodule in 3D.
Furthermore, adapting the value of $N$ per nodule type has the advantage of (1) balancing classes distribution in the presence of skewed distribution of classes by using a larger value of $N$ for underrepresented classes, and (2) using it as a kind of data augmentation, in which many different views of the same object are extracted.

The parameter $d$ defines the scale at which patches are considered.
Using multiple values of $d$ allows to crop triplets of patches with information that range from local content to more global context of nodule appearance.
In order to train the proposed deep learning system, we extracted triplets of patches at three different scales, namely $d=10, 20, 40$ mm and fed three streams of the network with three triplets at the same time.
This allows the network to focus both on the local appearance of a nodule (10 mm), where small structures like the solid core can be analyzed, and on more global context (40 mm), in which structures like the fissure can be recognized.
Before feeding the network, each patch was rescaled to a fixed size of 64$\times$64 pixels using bicubic interpolation and the pixel intensity $I_{HU} \in [-1200, 400]$ HU was rescaled to $I_{norm} \in [0, 1]$ by applying the transformation $I_{norm} = \frac{I_{HU} + 1200}{1600}$.

\subsection*{Deep learning network}

\subsubsection*{Network design}
The architecture of the used deep learning system is depicted in Figure \ref{fig:deeplearning}(b).
The system consists of nine streams of ConvNets, grouped into three sets of three streams.
Each set of streams is fed with a triplet of orthogonal patches extracted at the same scale.
Different sets of streams process triplets of orthogonal patches with exactly the same orientation in the CT scan, but at different scales.
Each stream of the set is fed with one patch from a triplet of orthogonal patches.
The 2D input patch is then processed by a series of convolutional and pooling layers, with one last fully-connected layer.
The size of each patch is 64$\times$64 pixels, which covers a size of $\approx$ 40 mm at the used in-plane resolution of 0.67 mm/px.

In order to define the optimal architecture for each stream, we followed the VGG-net  approach proposed in \cite{Simo14}.
We set a fixed size of convolutional kernels to 3$\times$3 px and used 32 filters in the initial layer.
Similarly to \cite{Simo14}, we added pairs of convolutional and max-pooling layers, keeping a fixed filter size of 3$\times$3 and doubling the number of filters in convolutional layers after each max-pooling, as long as the performance on the validation set were improving.
We slightly deviated from the fixed procedure of \cite{Simo14} by increasing the filter size in the first convolutional layer to $5\times5$ and by using 2 layers of 64 filters in cascade before the second max-pooling layer, since this configuration showed to perform slightly better than the standard one.
The described architecture represents one of the three streams used in a set, which we define as \emph{multi-stream} network.
All the parameters of the network are shared across the three streams in the same  multi-stream network.
It is worth noting that a multi-stream network processes triplets of 2D patches extracted with the same resolution $d$.

We used three scales with patch size of 10 mm, 20 mm and 40 mm, respectively, and for each scale we trained a multi-stream network.
Each multi-stream network has the same architecture, but parameters are optimized independently at each scale.
The multi-stream networks at different scales are finally merged in a final fully-connected layer (see Figure \ref{fig:deeplearning}(b)).
The final soft-max layer has six neurons, which produce the probability for the six considered classes.
We implemented the network using Theano\cite{Bast12}.

\subsubsection*{Training}
We trained the proposed multi-stream multi-scale convolutional network with data from the MILD trial.
For training purposes, we split the dataset into two parts, a \emph{training} set containing 75\% of the data, and a \emph{validation} set, containing the remaining 25\% of the data.
We defined the two data sets without any overlap of patients or nodules across the sets and distributing all nodule types in the two sets based on the same proportion 75\%-25\%.
The statistics of the two data sets are reported in Table \ref{tab:data}.

For training purposes, for each nodule, three triplets of patches were extracted.
Each triplet was extracted at a given scale by setting the values $d_1 = 10$ mm, $d_2 = 20$ mm and $d_3 = 40$ mm for the streams 1,2, and 3 respectively.
Since the distribution of nodule types were skewed, we adapted the number of angles $N$ per nodule type.
In order to set the proper value for $N$, we decided to initially extract 5,000 training samples per nodule class.
Specific values for $N$ for each class are reported in Table \ref{tab:data}.
Adapting the value of $N$ per nodule type produced 30,000 training samples.
We further augmented the size of the training data set by adding three shifted versions of each training sample.
Data augmentation was therefore done by randomly shifting the position $\mathbf{q}$ of the center of mass of the nodule to $\mathbf{q}_{shift} = [x + \delta_x, y+\delta_y, z+\delta_z]$, where $(\delta_x,\delta_y,\delta_z)$ were drawn from a normal distribution with mean value $\mu$= 0 and standard deviation $\sigma = \frac{1}{3\sqrt{3}}$, which ensures shifting within a sphere of radius 1 mm centered on $\mathbf{q}$.
Finally, each patch of the triplet and its shifted version were flipped along the vertical, the horizontal axis, and a combination of the two axes.
As a result, 16 different views of each nodule sample were included in the training set, which resulted in approximately 500,000 training samples.

In order to train the ConvNet, we initialized the parameters according to the method in \cite{Glor10} and trained using stochastic gradient descent, minimizing the categorical cross-entropy loss.
During optimization, we set an initial learning rate $\eta=10^{-3}$ and decreased it by a factor 3 every 50 epochs.
The parameters of the network were updated using the ADAM algorithm \cite{King15}.
We set the batch size to 256 and used dropout\cite{Kriz12} with a probability of 0.5 in the last fully-connected layer.
Additionally, L2 normalization was used, with a weight decay parameter of $10^{-6}$.
We empirically noticed that the training converges after $\approx$ 200 epochs.

\subsubsection*{Prediction}
Given an input sample $\mathbf{x}$, consisting of a set of triplets extracted at multiple scales, the trained  architecture is able to predict a probability $P_k(\mathbf{x})$ for each considered nodule type class $k$.
Since one set of triplets is extracted for a given angle $\theta$, the prediction also depends on the angle $\theta$.
Therefore, the input triplet for a given nodule can be written as a function of $\theta$, namely $\mathbf{x}_{\theta}$. 
In order to classify a given nodule, the prediction becomes a function of the parameter $\theta$ as well, which we can write as $P(\mathbf{x}_{\theta})$.
The final prediction is obtained as a combination of the $N$ predictions obtained by varying the parameter $\theta$.
The adopted combination strategy consisted in averaging the per-class probability, and finally assigning the nodule the label $y = argmax_k(\frac{1}{N}\sum_{i=1}^N P_k(\mathbf{x}_{\theta_i}))$.
This prediction strategy was applied both during training to assess the performance of the network on the validation set, and during the final evaluation on the DLCST data set.
For validation purpose, after each epoch, all nodules in the validation set were tested and performance was assessed.
For this purpose, 30 samples per nodule were extracted (\mbox{$N$ = 30}), meaning that patches at rotation st
 of 6$^{\circ}$ were taken.
At each iteration, nodule type was predicted using the proposed combination of predictions, and quantitative performance parameters were computed.
Since the distribution of nodule types in the validation set is skewed (see Table\ref{tab:data}), we considered the F-measure per class instead of the commonly used accuracy, since the F-measure is less sensitive to skewed distributions.
Based on this, during training we maximized the mean F-measure across classes.
For the final evaluation on DLCST data, the same settings using $N$=30 was used, and the results for the three considered architectures reported in Table \ref{tab:kappa} and Table \ref{tab:comparison} were obtained.

\subsection*{Nodule classification using Support Vector Machines}
In this section, we describe the details of the experiments based on classical machine learning approaches, where we used two different sets of features.
The first set consists of features based on the \emph{intensity} of pixels in 2D patches.
The second set consists of features automatically learned from raw data in an \emph{unsupervised} fashion, using the K-means algorithm.

\subsubsection*{Intensity features}
The first set of features consists of the raw pixel intensity (HU values) extracted from 2D patches.
Given a patch of size 64$\times$64 px, we extracted a feature vector by vectorizing the values of pixel intensities in the patch.
In this way, each patch had a 4,096-dimension feature vector.
We built a training set by considering all the nodules used to train the methods based on deep learning, balancing samples across classes using the coefficients reported in Table \ref{tab:data}.
We used the training set to train a linear Support Vector Machines (SVM) classifier.
Data were normalized prior to training to have zero mean and unit variance, and the one-vs-one strategy was used to deal with the multi-class problem.
After training, we applied the classifier to the $test_{ALL}$ dataset, which contains 634 nodules.
As done for the evaluation of deep learning approaches, 30 patches per nodules were considered at test time, which were all classified using the trained SVM classifier.
Finally, majority voting of the predicted labels was used to obtain the final prediction of nodule type.

\subsubsection*{Unsupervised features}
The approach used to learn a representation of pulmonary nodules in an automatic unsupervised fashion is based on the work of Coates et al. \cite{Coat11}.
The original method presented in \cite{Coat11} was developed based on the CIFAR10 dataset, which contains RGB images of 32$\times$32 px.
Since the size of the patches used in this paper is 64$\times$64 px, in order to apply the method in \cite{Coat11} to our data we doubled the receptive field size, which we set to 12 px, and set the number of centroids to 1,600, which gave a feature space of 6,400 dimensions.
We kept the rest of parameters of the algorithm at their default value.
As done for the experiment using intensity features and linear SVM, at test time we classified 30 samples per nodule and considered the label given by the majority voting on the predicted labels as the final prediction of nodule type.
\newline\newline


\section*{Acknowledgements}
This project was funded by a research grant from the Netherlands Organization for Scientific Research, project number 639.023.207. The MILD project was supported by grants from the Italian Association for Cancer Research (AIRC): IG research grant  11991 and the special program Innovative Tools for Cancer Risk Assessment and early Diagnosis, 5 1000, No.12162; Italian Ministry of Health (RF- 2010).
The authors would like to thank NVIDIA Corporation for the donation of a GeForce GTX Titan X graphics card used  in the experiments.

\section*{Author contributions statement}
F.C. conceived and conducted the experiments, analysed the results and wrote the manuscript. K.C. trained students for annotating training data, reviewed training set annotations and took part in the observer study. S.v.R. reviewed training set annotations. A.S. and P.G. assisted in the technical development of the deep learning system. C.J. assisted in data selection. E.T.S. reviewed training data annotations and took part in the observer study. C.S.P. took part in the observer study. M.W. provided data for the evaluation of the method. A.M. and U.P. provided data for the training of the method.
M.P. and B.v.G. designed and directed the study. All authors reviewed the manuscript.

\section*{Additional information}

\textbf{Competing financial interests}.  
Colin Jacobs received a research grant from MeVis Medical Solutions AG, Bremen, Germany.
Bram van Ginneken receives research support from MeVis Medical Solutions and is co-founder and stockholder of Thirona. 

\end{document}